\documentclass[10pt,twocolumn,letterpaper]{article}

\usepackage{3dv}
\usepackage{times}
\usepackage{epsfig}
\usepackage{graphicx}
\usepackage{amsmath}
\usepackage{amssymb}
\usepackage{array,multirow}
\usepackage{enumitem}
\usepackage{fancyhdr}

\usepackage[pagebackref=true,breaklinks=true,colorlinks,bookmarks=false]{hyperref}

\newcommand\phantomarrow[2]{%
  \setbox0=\hbox{$\displaystyle #1\to$}%
  \hbox to \wd0{%
    $#2\mapstochar
     \cleaders\hbox{$\mkern-1mu\relbar\mkern-3mu$}\hfill
     \mkern-7mu\rightarrow$}%
  \,}
\newcommand{\bm}[1]{\mathbf{#1}}

\newcommand{\subheading}[1]{\textbf{#1}.}

\fancypagestyle{title}{%
  \setlength{\headheight}{22pt}%
  \fancyhf{}
  \fancyhead[L]{\small© 2021 IEEE. Personal use of this material is permitted.
  Permission from IEEE must be obtained for all other uses, in any current or future media, including reprinting/republishing this material for advertising or promotional purposes,
  creating new collective works, for resale or redistribution to servers or lists, or reuse of any copyrighted component of this work in other works}
}%

\threedvfinalcopy 

\def\threedvPaperID{170} 

\ifthreedvfinal\fi
\begin{document}

\title{AIR-Nets: An Attention-Based Framework for \\ Locally Conditioned Implicit Representations}




\author{Simon Giebenhain 
\qquad \qquad 
Bastian Goldlücke\\
University of Konstanz, Germany\\
{\tt\small \{simon.giebenhain, bastian.goldluecke\}@uni-konstanz.de}
}

\maketitle
\thispagestyle{title}

\begin{abstract}
   This paper introduces Attentive Implicit Representation Networks (AIR-Nets), a simple, but highly effective architecture for 3D reconstruction from point clouds. Since representing 3D shapes in a local and modular fashion increases generalization and reconstruction quality,
   AIR-Nets encode an input point cloud into a set of local latent vectors anchored in 3D space, which locally describe the object's geometry, as well as a global latent description, enforcing global consistency. Our model is the first grid-free, encoder-based  approach that locally describes an implicit function.
   The vector attention mechanism from \cite{PointTransformer} serves as main point cloud processing module, and allows for permutation invariance and translation equivariance. 
   When queried with a 3D coordinate, our decoder gathers information from the global and nearby local latent vectors in order to predict an occupancy value.
   Experiments on the ShapeNet dataset~\cite{ShapeNet} show that AIR-Nets significantly outperform previous state-of-the-art encoder-based, implicit shape learning methods and especially dominate in the sparse setting. Furthermore, our model generalizes well to the FAUST dataset~\cite{FAUST} in a zero-shot setting.
   Finally, since AIR-Nets use a sparse latent representation and follow a simple operating scheme, the model offers several exiting avenues for future work. Our code is available at  \small{\url{https://github.com/SimonGiebenhain/AIR-Nets}}.
\end{abstract}

\vspace{-15pt}

\section{Introduction}
Humans naturally possess a remarkable ability to perceive their environments in 3D. A few glimpses and head movements suffice to grasp the details of an object and understand its position within a scene. For machines such tasks remain difficult to solve.


For deep learning (DL) methods, choosing a suitable representation of 3D objects is defining large parts of the challenges to be solved, especially when generating 3D objects. For example, voxel grids provide great regularity and access to rich research results of convolutional neural networks (CNNs), but often have to be restricted in their resolution. Other representations like point clouds and meshes live in a much sparser domain, but come with their own problems: while point clouds usually require an extreme number of points to describe detailed shapes, generating meshes with DL approaches is complicated, due to the discrete, combinatorial intricacies of generating face connectivities.

In order to avoid such problems, researches recently explored implicit neural representations \cite{DeepSDF, OccNets, IM-Net}, a fully continuous alternative for watertight objects, that is easily implemented and scales well on GPUs. The most common choices for the implicit function are signed distance functions (SDFs) and occupancy functions (zero outside and one inside of the object). In both cases a neural network approximates the implicit function of a watertight object, from which the 3D surface can be extracted by finding the zero level-set or decision boundary, respectively. 
While earlier work~\cite{OccNets, DeepSDF, IM-Net} always used a single latent vector to condition the implicit function on a specific object, recently research shifted to representing objects locally~\cite{if-net, ConvOccNets, Jiang_2020_CVPR, Chabra2020DeepLS, PatchNets}, resulting in greater representational power and higher generalization capabilities. Of these methods,~\cite{ConvOccNets, if-net} are the only ones that use an encoder to infer the latent representation. The other models at least partially rely on running an optimization algorithm to infer an encoding,
as inspired by~\cite{DeepSDF}. 

\begin{figure*}[ht!]
\begin{center}
\def\svgwidth{\linewidth}
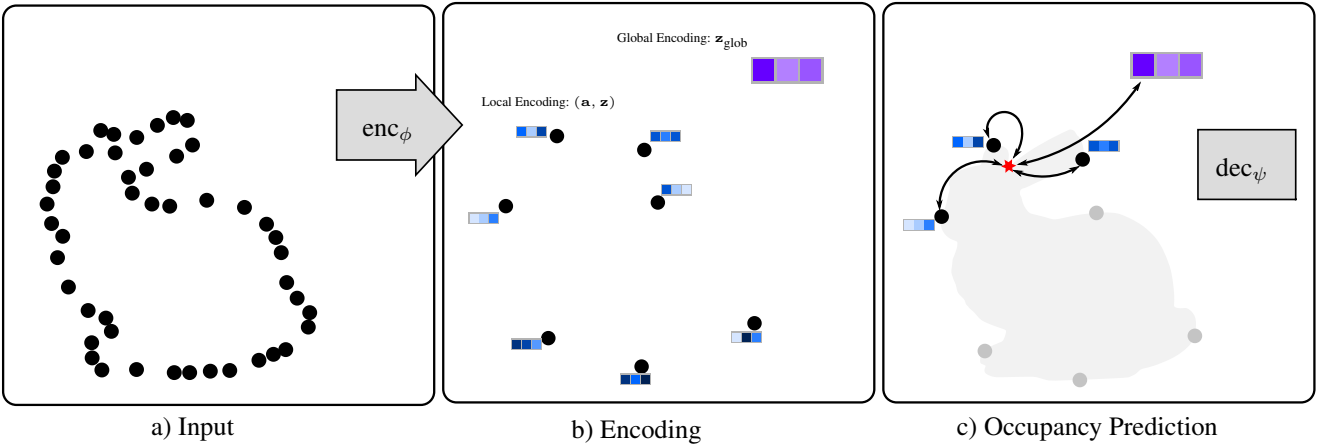
\end{center}
   \caption{\textbf{Model Overview:} This figure illustrates the operating scheme of AIR-Nets simplified to 2D. The input point cloud (left) is encoded into a global latent vector (purple) and a much sparser point cloud, where a local latent vector (blue) is attached to each point (middle). When queried with a spatial coordinate (red star), the decoder network locally extracts latent information using an attention mechanism, where the global descriptor acts as query and the local ones act as keys and values, to predict an occupancy value (right).}
\label{fig:overview}
\end{figure*}

In this paper, we propose \emph{Attentive Implicit Representation Networks} (AIR-Nets), which are, to the best of our knowledge,
the first encoder-based local implicit shape learning model that purely operates in the point cloud domain.
AIR-Nets utilize the vector attention mechanism proposed in \cite{PointTransformer} to encode input points into a sparse set of locally anchored latent vectors and a single global latent vector. By representing the geometric structure of objects as a composition of local descriptors, AIR-Nets exhibit highly detailed reconstructions and great generalization capabilities. The occupancy function is modeled by our proposed attentive decoder, which aggregates information from the global and nearby local latent vectors. These extracted features are finally transformed into occupancy predictions by a simple feed-forward network (FFN). The resulting architecture is permutation invariant and translation equivariant, see figure \ref{fig:overview} for a schematic illustration.
Therefore, compared to the grid-based methods of \cite{if-net, ConvOccNets}, which represent shapes with features drawn from the dense voxel grid of a 3D CNN, AIR-Nets avoid any discretization, while representing objects in a more compressed fashion. This allows for easy adoptions in downstream tasks, new domains or translation to a probabilistic generative setting. 


In summary, our contributions are the following:
\begin{itemize}[noitemsep]
    \item We propose the AIR-Net framework, a highly effective, purely point cloud based architecture for local, implicit shape representations, that modularly encodes objects into a set of locally anchored latent vectors.
	\item We propose a novel attention-based set abstraction method that outperforms the common maxpooling based downsampling.
	\item Furthermore, we propose an attentive decoder in conjunction with our latent shape representation and show the importance of an expressive decoder compared to a simple interpolation based decoder.
	\item AIR-Nets significantly outperform state-of-the-art implicit representations methods in 3D reconstruction from point clouds and generalize well from the ShapeNet to the FAUST dataset in a zero-shot setting.
\end{itemize}

\section{Related Work}

Designing DL methods for geometric objects heavily depends on the chosen representation for 3D shapes. We therefore present a brief overview of DL architectures for geometric data processing in section \ref{geometric_data_processing}, while models for shape generation and 3D reconstruction are discussed in section \ref{shape_generation}.


\subsection{Geometric Data Processing}
\label{geometric_data_processing}
This section provides a brief overview of methods for discriminative tasks on geometric data. 
For a great theoretical approach to geometric DL and a more detailed overview of this rapidly growing field we recommend \cite{bronstein2021geometric}. 

\subheading{Voxel Processing}
Representing 3D objects as voxel grids and generalizing tools from 2D CNNs is possibly the most straight forward way to process 3D shapes, \eg \cite{brock2016generative, OctTree}. 
Despite their cubic memory demands, 3D CNNs are frequently adopted, \eg in the closely related works \cite{OccNets, ConvOccNets, if-net} or in 3D object detection \cite{PVRCNN}, which uses sparse convolutions \cite{sparseConv} to lessen the memory footprint.

\subheading{Point Cloud Processing} 
Designing DL models for point clouds is complicated, since they have to be permutation invariant and point clouds often exhibit very irregular structure and densities. Early approaches process points individually before applying permutation invariant pooling operations~\cite{PointNet, PointNet++}. More recent approaches utilize graph convolutional networks (GCN), originally introduced in~\cite{bruna_spectral,  Defferrard, KipfWelling, Gilmer_MPNN}. For example~\cite{DGCNN, PointWeb, GridGCN} use GCNs by locally connecting points in the point cloud.

Others define continuous convolution operations around points, \eg \cite{PCCN, PointConv, KPConv}. More models are surveyed in \cite{surveyDeep3D}.

Inspired by the recent success of Transformers~\cite{vaswani2017attention}, their natural permutation invariance and ability to work with a wide variety of data structures, \eg sequences~\cite{vaswani2017attention, bioinformatics_rives}, images \cite{ViT} or abstract objects~\cite{DETR}, recent work builds upon attention mechanisms to process point clouds~\cite{PointTransformer, PointCloudTransformer, engelPointTransformer}. Our approach leverages the ideas from~\cite{PointTransformer}, which uses the generalized vector attention mechanism from~\cite{expSA}, instead of the classical scalar-valued dot-product attention from~\cite{vaswani2017attention}.
Note that attention mechanisms can be thought of a special kind of graph convolution operation, as pointed out in~\cite{bronstein2021geometric}, i.e. full attention operates on the fully connected graph, while the local attention of \cite{PointTransformer} operates on the $k$ nearest neighbor ($k$NN) graph.
Finally, we want to highlight some approaches to obtain SE(3)-equivariant processing modules~\cite{fuchs2020se3transformers, hutchinson2020lietransformer, EnEGNN}.

\subheading{Mesh Processing} 
Methods for meshes are closely related to  GCNs, which are, however, insensitive to the local geometry of the mesh. Therefore, several different approaches to obtain geometry aware convolutions exist~\cite{Masci2015GeodesicCN, Monti2017GeometricDL, GEM}. We refer to~\cite{bronstein2021geometric} for a detailed overview.

\subsection{Shape Generation and Reconstruction}
\label{shape_generation}
Contrary to the previous section, this section is solely concerned with models that produce a 3D shape as output, \eg generative or 3D reconstruction models. As point clouds do not unambiguously describe a surface, models generating point clouds are not discussed here.

\subheading{Voxel Grids}
Generating high resolution voxel grids is less suitable, due to a high memory footprint. While it is possible to design more efficient octree-based 3D CNN architectures \cite{OctTree, haene_hsp}, these methods are still limited by a resolution of $256^3$ and are complicated to implement.

\subheading{Meshes} 
Usually mesh generation is limited to a single topology of a \emph{template} mesh.
Therefore the generation task condenses to predicting the location of all vertices. Such models \cite{SpiralNet++, CoMA} achieve impressive results in this less general scenario. Template-free mesh generation is less common, since predicting a variable number of vertices and their adjacencies is quite complicated. \cite{PolyGen} build on Transformers \cite{vaswani2017attention}, by viewing the problem as a sequence generation problem and auto-regressively predicting all vertices and then all $n\text{-gon}$ faces. A similar approach for general graphs is proposed in \cite{liao2019gran}.

\subheading{Implicit Continuous Functions}
To avoid some of the above mention complications, continuous implicit functions like the (truncated) SDF and occupancy function have become a rapidly growing field of research for 3D representations of \emph{watertight} objects.
Due to their continuity, implicit functions are well-suited for DL and can be easily implemented as a simple FFN. Such models mainly differ in two aspects. First, some models use an encoder network to determine the latent representation~\cite{OccNets, IM-Net}, while others are \emph{encoder-less}~\cite{DeepSDF, metaSDF, im_geom_reg}, which eliminates the need to design an encoder for geometric observations such as point clouds. Instead they directly optimize for the latent representation of each object, causing long inference time for previously unseen objects.

The second major difference lies in the conditioning of the implicit function on a specific object. While early approaches use a single latent vector to represent a shape~\cite{OccNets, IM-Net, DeepSDF}, recent models use multiple local latent descriptors, resulting in high-quality reconstructions and better generalization. The encoder-based approaches~\cite{if-net, ConvOccNets} represent the implicit function locally using features drawn from the voxel grid of a CNN. Compared to these grid-based approaches, our model purely operates in the point cloud domain, avoiding discretization losses and quadratic/cubic scaling with the resolution, and resulting in a much sparser and more compressed encoding. 

Several approaches to encoder-less local representations exist~\cite{PatchNets, Jiang_2020_CVPR, Chabra2020DeepLS}. These methods, however, cannot effectively reconstruct large regions devoid of any observations.

The models in~\cite{LDIF} and~\cite{PatchNets} are similar to ours, since they also use a sparse set of spatially anchored latent vectors, but construct their implicit function differently, by interpolating between multiple implicit functions using 3D Gaussians. Our model uses a computationally more expressive and slightly less interpretable procedure to parameterize the implicit function.

Using a slightly different approach, \cite{NDF} models an unsigned distance field, relaxing the assumption of watertight objects, but rendering the extraction of the zero level set impossible. Instead very dense surface point clouds and rendered images of the surface can be generated.

\section{Attentive Implicit Representation Networks}


AIR-Nets offer an encoder-based implicit shape learning framework, which finds a compromise
between conditioning the implicit function on a single latent vector \cite{OccNets, IM-Net} and a dense voxel grid of features \cite{if-net, ConvOccNets}. It does so by encoding an input point cloud into a sparse set of $M$ locally anchored latent vectors and a global latent vector. Therefore \hbox{AIR-Nets} are encouraged to understand objects as a composition of local shapes, leading to better generalization capabilities and a higher capacity to reconstructed rich details. Figure \ref{fig:overview} illustrates the operating scheme of AIR-Nets. 

From a high-level point of view, the encoder
\begin{equation}
\operatorname{enc}_{\phi} \colon \mathbb{R}^{N\times (3 + d_0)} \to \mathbb{R}^{M\times (3+d)} \times \mathbb{R}^d \mathbb,
\end{equation}
maps the input point cloud $P \in \mathbb{R}^{N \times 3}$ and features $F \in \mathbb{R}^{N \times d_0}$ to anchors $\mathbf{a}$, local latent vectors $\mathbf{z}$ and the global latent vector $\mathbf{z}_{\text{glob}}$. The complete encoder architecture is described in section \ref{sec:encoder}, after reviewing the vector attention mechanism of \cite{PointTransformer} in section \ref{sec:background}.

The implicit function is modeled by the decoder 
\begin{equation}
\operatorname{dec}_{\psi} \colon \mathbb{R}^{M\times (3 + d)} \times \mathbb{R}^d \times \mathbb{R}^3 \to [0, 1],
\end{equation}
which takes queried coordinates $(x,y,z)\in \mathbb{R}^3$ alongside the encoding as input, and predicts the probability for $(x,y,z)$ to be inside of the observed watertight object. The decoder aggregates information from nearby local latent vectors using vector cross attention.
Finally, a simple FFN maps the locally aggregated information to an occupancy prediction. More details about the decoder are presented in section \ref{sec:decoder}. 

\subsection{Background: Point Transformer
\cite{PointTransformer}}
\label{sec:background}

The vector attention mechanism of \cite{PointTransformer} serves as core processing module throughout the complete AIR-Net architecture. Therefore, this section formulates their vector attention mechanism in the more general form of a cross attention operator. The vector cross attention (VCA) operator
\begin{equation}
	\operatorname{VCA} \colon \mathbb{R}^{N_q \times (3+d_q)} \times \mathbb{R}^{N_{kv}\times (3+d_{kv})} \to \mathbb{R}^{N_q \times d}
	\label{VCA_eq}
\end{equation}
maps a \emph{query} point cloud $P_q \in \mathbb{R}^{N_q \times 3}$ and features $X_q \in \mathbb{R}^{N_q \times d_q}$  and a \emph{key-value} point cloud $P_{kv}\in \mathbb{R}^{N_{kv}\times 3}$ and features $X_{kv} \in \mathbb{R}^{N_{kv}\times d_{kv}}$ to output features $X^{\prime} \in \mathbb{R}^{N_q \times d}$. From a high level perspective, attention mechanisms can be thought of as a differentiable information routing schemes, where queries describe the sought for information among the key-value tokens. 

Internally, queries
$Q = (P_q, F_q) = (P_q, X_qW_q)$ and  keys $K = (P_{kv}, F_k) = (P_{kv}, X_{kv}W_k)$ are derived using linear transformations  $W_q \in \mathbb{R}^{d_q \times d}$ and $W_k \in \mathbb{R}^{d_{kv}\times d}$ on the features. Information around the $i\text{-th}$ query is then aggregated from its neighborhood $\mathcal{N}_i$ (i.e. $k\text{NN}$) using
\begin{equation}
	X_i^{\prime} = \sum_{j\in \mathcal{N}_i}n\left(s\left(Q_i, K_j\right)\right) \odot V_{ij},
	\label{vector_attention_eq}
\end{equation}
where $n$ denotes the channel-wise softmax function, $\odot$ the Hadamard product of two vectors and 
\begin{equation}
V_{ij} = F_{kv_j}W_{v} + \delta(P_{q_i} - P_{kv_j})
\label{eq:values}
\end{equation}
the values. Note that defining the values as such, extends the standard formulation of the attention mechanism from \cite{vaswani} due to the dependence on $i$. 
Including positional information modelled by a two-layer multilayer perceptron (MLP) with a single ReLU activation $\delta \colon \mathbb{R}^3 \to \mathbb{R}^{d}$ was empirically found to be beneficial \cite{PointTransformer}. The original motivation for $\delta$ lies, however, in the similarity function 
\begin{equation}
s(Q_i, K_j) = \gamma \left(F_{q_i} - F_{k_j} + \delta(P_{q_i} - P_{kv_j})\right),
\label{eq:similarity_ptb}
\end{equation}
which calculates vector valued similarities between queries and keys using $\delta$ and another two-layer MLP $\gamma$. Using this more general $\operatorname{VCA}$ operator, vector self attention (VSA), as used in the Point Transformer, can be finally defined as 
\begin{equation}
    \operatorname{VSA}(X) := \operatorname{VCA}(X, X).
\end{equation}
We refer to the supplementary for a formulation of $\operatorname{VCA}$ and $\operatorname{VSA}$ in the language of graph neural networks.

\subsection{Encoder}
\label{sec:encoder}

\begin{figure*}[t]
\begin{center}
\def\svgwidth{0.85\linewidth}
 		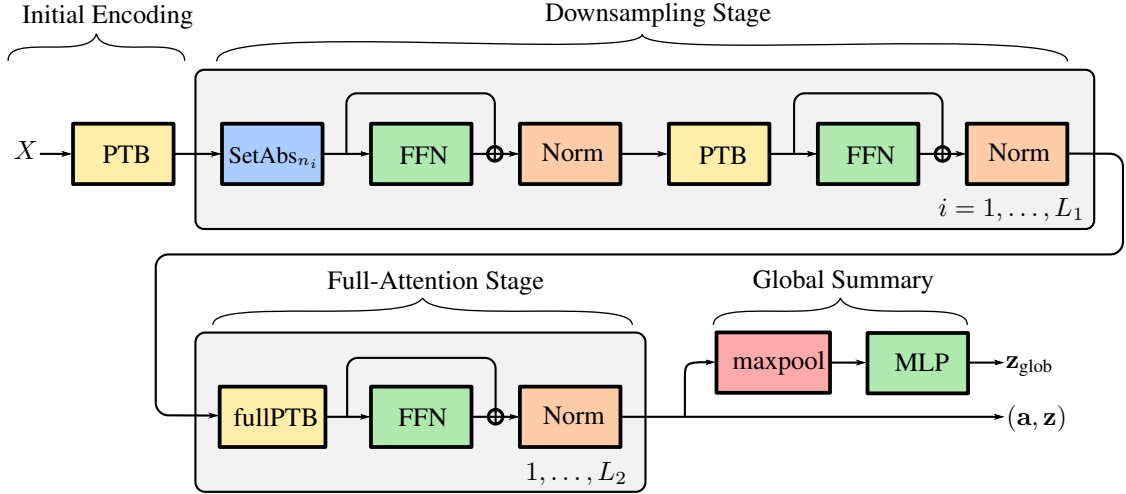
\end{center}
   \caption{\textbf{Encoder architecture:} After an initial encoding, the point cloud is repeatedly downsampled using $L_1$ SetAbs modules. To preserve more information, an additional PTB is added before the next downsampling step. Each downsampling layer also includes a FFN, skip connection and normalization layer after each block involving attention, as proposed in \cite{vaswani2017attention}. In the same style, $L_2$ full-attention PTBs are used to promote global information exchange, producing local descriptors $\bm{z}$ and anchors $\bm{a}$. The global latent vector $\bm{z}_{\text{glob}}$ is derived using maxpooling. All normalization layers are realized as BatchNorm layers \cite{BatchNorm}.}
\label{fig:encoder}
\end{figure*}

This section describes the encoder architecture, as illustrated in figure \ref{fig:encoder}. The two main modules of the encoder are the \emph{point transformer block} (PTB) and \emph{attentive set abstraction module} (SetAbs), which we describe later. Both are based on the VSA and VCA modules, as defined above. The point transformer block 
\begin{equation}
    \operatorname{PTB}(X) := \operatorname{BN}(X + \operatorname{VSA}(X))
\end{equation}
simply encapsulates VSA, a residual connection and a BatchNorm (BN) layer \cite{BatchNorm}. Note that the point's positions remain unaltered. We form local neighborhoods using $k$NN with $k_{\text{enc}}=16$, to impose an effective inductive bias and limit memory consumption.

\subheading{Translation Equivariance}
Here, we want to stress that the whole network is translation equivariant, since both VSA and VCA modules only use relative positions. Therefore the computation of features is invariant, which are, however, spatially anchored and hence become equivariant.
In order to preserve this equivariance it is important to not include the input coordinates as features in the initial encoding PTB. If the input point cloud does not carry any other features besides the spatial coordinates (i.e. $F$ is empty), eq.  (\ref{eq:values}) and (\ref{eq:similarity_ptb}) have to be modified such that only the $\delta\text{-terms}$ remain.

\begin{figure}[h!]
\begin{center}
\def\svgwidth{\linewidth }
			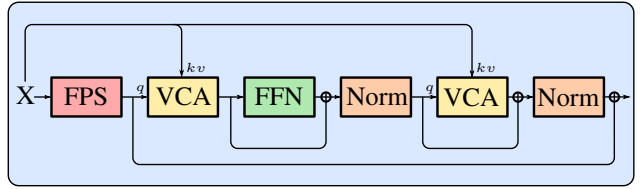
\end{center}
   \caption{The proposed set abstraction module~$\operatorname{SetAbs}_n$ first selects $n$ central points using FPS. The central points act as queries in a VCA module that aggregates information from their neighborhoods of size~$k_{\text{enc}}=16$. A second round of attention lessens the dependence on the inital, random selection of the central point. Here FFN composes a 2-layer MLP, skip connection and BN layer.}
\label{fig:setAbs}
\end{figure}

\subheading{Downsampling}
In many learning tasks, downsampling is a crucial component to encourage abstract and robust reasoning and reduce the computational requirements.
AIR-Nets follow the typical downsampling scheme applied in most point cloud processing networks \cite{PointNet++, PointTransformer, PointCloudTransformer}. The prevalent procedure is to subselect points using farthest point sampling (FPS) as a simple heuristic and then build local neighborhoods around the selected points (from now on called \emph{central} points) using a grouping mechanism, like ball queries~\cite{PointNet++} or $k\text{NN}$. Finally, each neighborhood is abstracted into the new representative features for the central point's position using a permutation invariant function. Transforming point features independently with a small, shared MLP followed by maxpooling is the most common choice for the set abstraction step. 
We argue, however, that maxpooling is sub-optimal, since each point is treated equally, which becomes especially problematic in very diverse neighborhoods and in the presence of outliers. As an alternative to maxpooling, we propose to use two iterations of vector cross attention. The complete attentive set abstraction module is depicted in figure \ref{fig:setAbs} and adheres to the function signature
\begin{equation}
	\operatorname{SetAbs_{n^{\prime}}} \colon \mathbb{R}^{n \times (3+d)} \to \mathbb{R}^{n^{\prime} \times (3+d)},
\end{equation}
where $n^{\prime}$ specifies the desired output cardinality. The intuition behind the module is that the central point attends to its $k_{\text{enc}}$ neighbors (including itself) and builds a summary thereof. However, since this summary strongly depends on the (random) selection of the central point, a second round of attention refines the summary to lessen this bias. 

Using $\operatorname{SetAbs}$ as a downsampling module, the point cloud is repeatedly downsampled using $L_1$ downsampling layers with intermediate cardinalities $n_1, \dots , n_{L_1-1}$, until a cardinality of $n_{L_1}:=M$ is reached. In order to give the model an opportunity to compute more meaningful features before the next SetAbs, each downsampling layer also includes a regular PTB.

\subheading{Full Attention}
After the input is reduced to the desired cardinality $M$, $L_2$ full attention layers facilitate global information exchange (instead of $k$NN, all points are considered as neighbors). Gaining a global understanding of the object is important in order to exploit symmetries and supplement information to regions with little observations. The resulting point cloud and features serve as anchors $\mathbf{a}$ and local latent vectors $\mathbf{z}$ respectively.

Finally, the global summary $\mathbf{z}_{\text{glob}}$ is obtained using global maxpooling followed by a 2-layer MLP. 


\subsection{Decoder}
\label{sec:decoder}

Provided with an encoding $(\mathbf{a}, \mathbf{z}), \mathbf{z}_{\text{glob}}$ of a watertight object, the decoder
\begin{equation}
\operatorname{dec}_{(\mathbf{a}, \mathbf{z}), \mathbf{z}_{\text{glob}}} \colon \mathbb{R}^3 \to [0, 1], (x, y, z) \mapsto o
\end{equation}
models the occupancy function, which maps coordinates $(x,y,z)$ to the probability $o$ to be inside of the object.

Due to the compressed nature of the encoding, the relevant information for an occupancy prediction is not as regularly structured and not as readily available as in grid-based approaches \cite{ConvOccNets, if-net}. Consequently, we find that using an expressive enough decoder, capable of effectively extracting the relevant information in many different scenarios,
is crucial, as ablated in section \ref{sec:ablation_study}.
Therefore we propose to use vector cross attention to extract the relevant information
\begin{equation}
    \mathbf{z}_{\text{loc}} = \operatorname{VCA}(X_q, X_{kv}),
\end{equation}
where $X_q=((x,y,z), \mathbf{z}_{\text{glob}})$ is the single element query and $X_{kv}=(\mathbf{a}, \mathbf{z})$ constitutes key-value tokens. To further encourage a local encoding and reduce the runtime and memory complexity, the attention is limited to the $k_{\text{dec}}$ nearest local latent vectors. Using~$\bm{z}_{\text{glob}}$ as query provides context information, adapting the local information aggregation appropriately to the given scenario.
Furthermore, we find that additionally including $\mathbf{z}_{\text{glob}}$ in the key-values achieves slightly better results. For this token alone we remove the delta term from
eq.~(\ref{eq:values}), as it is free of any spatial location.

The final occupancy probability $o$ is predicted using the same FFN as in \cite{Niemeyer2020CVPR, ConvOccNets}.

\subsection{Training and Inference}
\label{sec:training}

The model parameters $\phi$ and $\psi$ are trained using a dataset of $T$ watertight shapes $S_i$, with inputs $X_i = (P_i, F_i)$ and ground truth $Y_i = (Q_i, O_i)$, where $Q_i$ denotes the 3D coordinates and $O_i$ the corresponding ground truth occupancy values. Note that in all of our experiments $P_i$ is the only observation (\ie $F_i$ does not carry any additional features).\\
We optimize the parameters $\phi$ and $\psi$ using the loss
\begin{equation}
    \mathcal{L} = \sum_{i\in I} \sum_{j \in J} L( \operatorname{dec}(\operatorname{enc}(X_i), Q_{i,j}), O_{i,j}),
\end{equation}
where $I$ is the mini-batch index, $J$ subselects $1.000$ ground truth points and $L(\hat{o}, o) = - [o \cdot \operatorname{log}(\hat{o}) + (1-o) \cdot \operatorname{log}(1- \hat{o})]$ is the binary cross-entropy loss. \\
During inference time, meshes are reconstructed using the MISE \cite{OccNets} algorithm, which is an efficient way of extracting the isosurface using the marching cubes algorithm \cite{marching_cubes}.

\section{Experiments}
\label{sec:experiments}
We conduct extensive experiments to demonstrate the representational power and generalization capability of AIR-Nets in 3D shape learning. In section \ref{sec:3d_rec} we evaluate our model on 3D shape reconstruction from point clouds using the ShapeNet \cite{ShapeNet} dataset and compare against a set of representative baseline models. Furthermore, we evaluate the zero-shot generalization capabilities of all these models by applying them to the FAUST \cite{FAUST} dataset in section \ref{sec:zero-shot}. Finally, we conduct controlled ablation studies in section \ref{sec:ablation_study} to validate several design choices of AIR-Nets.

\subsection{Datasets}
\label{sec:data_preparation}

We use all 13 object classes of the well-known ShapeNet dataset and use the common training split of \cite{choy}. Furthermore, we use two differently pre-processed versions of the dataset, $\mathcal{D}_A$ \cite{if-net} and $\mathcal{D}_B$ \cite{OccNets} and refer to the respective papers for more details. Note that therefore, the results cannot be directly compared across datasets.
The main difference of these datasets is, however, the spatial distribution of the supervision points $Q_i$. Dataset $\mathcal{D}_A$ samples $Q_i$ close to the boundary of the ground truth surface, 
in order to supervise more in the most crucial region.
Contrary $\mathcal{D}_B$ samples $Q_i$ uniformly in the unit cube. Furthermore, we follow \cite{if-net} and use $\mathcal{D}_A$ without measurement noise and follow \cite{OccNets, ConvOccNets} to use $\mathcal{D}_B$ with additive Gaussian measurement noise with a standard deviation of $0.005$.

Additionally we evaluate all models on the test set of the FAUST dataset without pre-processing the meshes.

\subsection{Baselines and Evaluation Metrics}
We compare against all state-of-the-art encoder-based implicit shape representation models.
We compare against \emph{Occupancy Networks} (ONets) \cite{OccNets}, as representative of models using a single global latent representation.
We additionally compare against \emph{Implicit Feature Networks} (IF-Nets) \cite{if-net} and \emph{Convolutional Occupancy Networks} (ConvONets) \cite{ConvOccNets}, as representatives for encoder-based local implicit representations. The former uses a 3D CNN to encode the input and for the latter we chose its best performing variant, which projects the 3D input points on the $xy$, $xz$ and $yz\text{-plane}$, and uses 2D CNNs.

The quality of the reconstructions is evaluated using the four most established metrics: the \emph{volumetric intersection over union} (IoU), the $L_1\textit{-Chamfer-Distance}$ ($L_1\text{-CD}$), the \emph{normal consistency} (NC) \cite{OccNets} and the \hbox{\emph{F-score}} \cite{F-score}.

\subsection{Implementation Details}
\label{sec:implementation_details}

We implemented our models using PyTorch \cite{pytorch} and trained all models using the Adam optimizer \cite{adam} and a batch size of 64.
For AIR-Nets the initial learning rate is set to~$5e^{-4}$ and decayed with a factor of~$0.2$ after every 200 epochs. For all baselines we used the official code release and training procedure. All models were trained until the validation loss ceased to improve. We select the epoch with the best validation loss for evaluation.

Unless stated otherwise the internal dimensions in all linear layers, the VCA and VSA modules is 256. The number of anchors is~$M=100$ and the number of downsampling layers is~$L_1=2$ for all experiments. When the number of input points~$N=300$, we set the intermediate cardinality of the downsampling stage to~$n_1 = 200$. When~$N=3000$, we use~$n_1 = 500$ instead.
The number of full attention layers is $L_2=3$. The neighborhood sizes are~$k_{\text{enc}}=16$ and~$k_{\text{dec}}=7$ unless stated otherwise. The VCA module in the decoder uses 200 dimensions to reduce memory demands. For the final FFN we use 5 layers and a hidden dimensionality of 128. Some of these hyperparameters as well as a runtime and memory analysis are provided in the supplementary. 

\subsection{3D Reconstruction from Point Clouds}
\label{sec:3d_rec}

In this section, AIR-Nets and the various baseline models are separately trained and evaluated on the 3D reconstruction task from point clouds on $\mathcal{D}_A$ and $\mathcal{D}_B$. 
Following \cite{if-net}, we additionally distinguish between a sparse and dense setting. In the sparse setting, inputs are composed of 300 points sampled uniformly on the ground truth object surface, while 3000 points are used for the dense setting.

While the sparse setting assesses the model's capabilities to complete geometric structures, the dense setting tests the representational capacity and ability to preserve fine details.

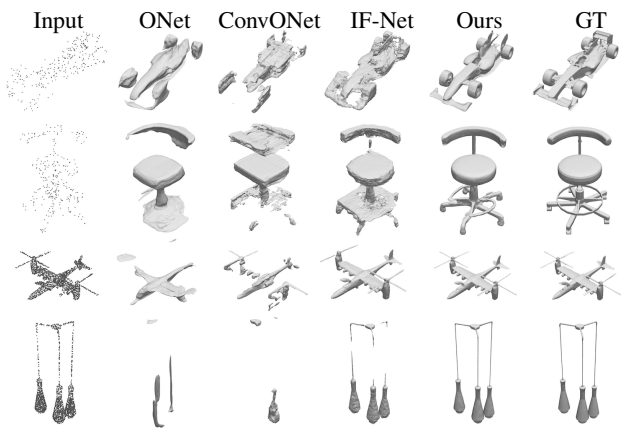
\begin{figure}[t]
    \begin{picture}(100,150)
        \put(0,0){\includegraphics[width=\linewidth]{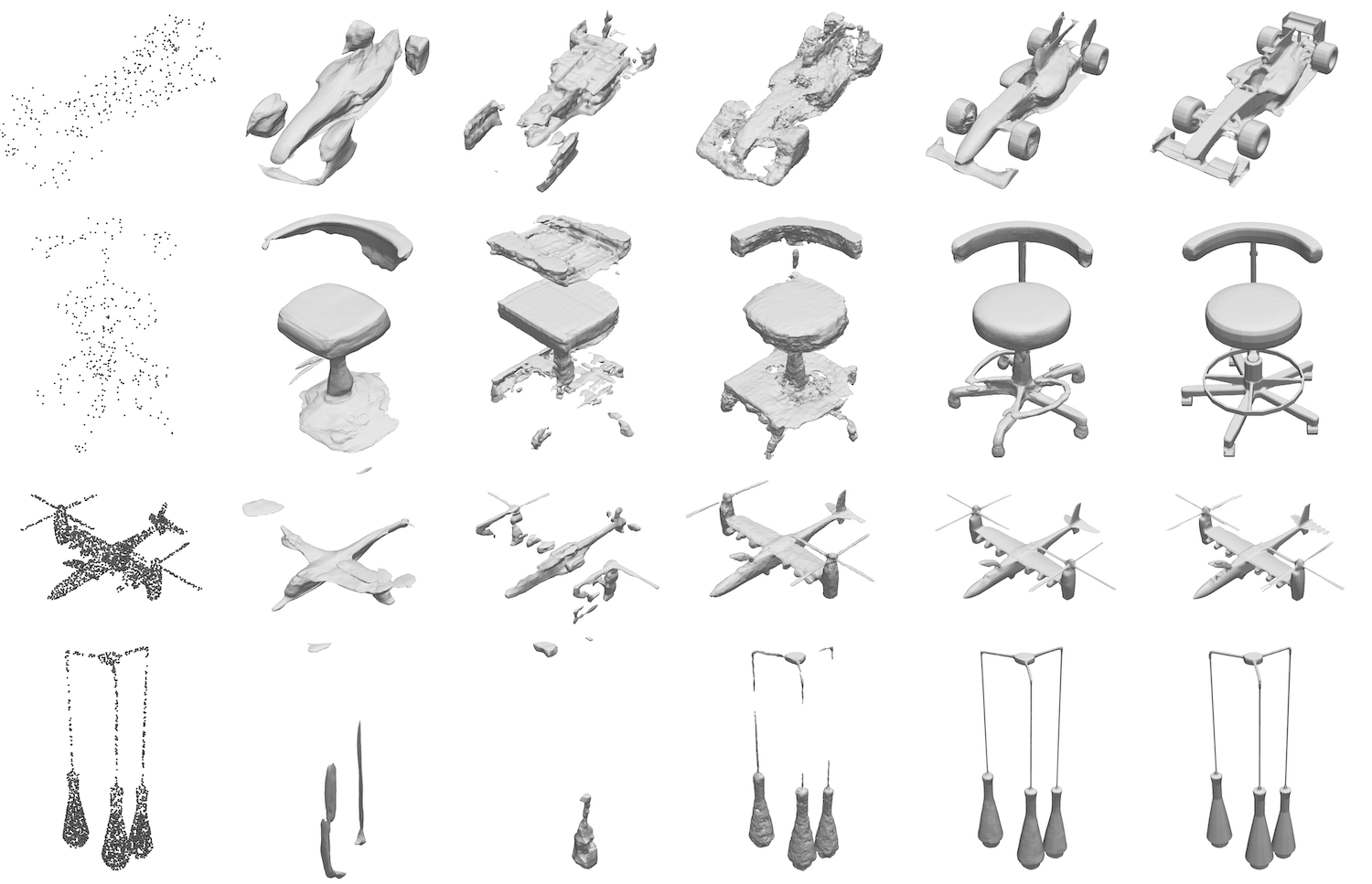}}
        \put(10,152){\small{Input}}
        \put(50,152){\small{ONet}}
        \put(80,152){\small{ConvONet}}
        \put(130,152){\small{IF-Net}}
        \put(170,152){\small{Ours}}
        \put(215,152){\small{GT}}
    \end{picture}
    \caption{\textbf{3D Reconstructions on $\mathcal{D}_A$}: Qualitative comparison on less common objects. The top two rows show the sparse setting, while the bottom two show results in the dense setting. Best viewed digitally and zoomed-in.
    }
    \label{fig:recs_DA}
\end{figure}


\subsubsection{Noise-free Observations from $\mathcal{D}_A$}
\label{sec:rec_d_a}

\begin{table}[t]
	\begin{center}
		\begin{tabular}{|c|l|c|c|c|c|} \hline
			&\multicolumn{1}{l|}{Model}&
			\multicolumn{1}{c|}{IoU$\uparrow$}&
			\multicolumn{1}{c|}{$L_1\text{-CD}$$\downarrow$}&
			\multicolumn{1}{c|}{\small{NC$\uparrow$}}&
			\multicolumn{1}{c|}{\small{F-Score$\uparrow$}} \\ \hline
			\parbox[t]{2mm}{\multirow{4}{*}{\rotatebox[origin=c]{90}{\footnotesize{Sparse Input}}}}
			& \small{ONet}     &         0.724 &          0.0094 &         0.878 &          0.808\\ 
			& \small{ConvONet} &         0.723 &          0.0084 &         0.876 &          0.841\\ 
			& \small{IF-Net}   &         0.772 &          0.0060 &         0.885 &          0.885\\ 
			& \small{Ours}     & \textbf{0.913} & \textbf{0.0032}& \textbf{0.95}  & \textbf{0.964}\\
			 \hline
			 \parbox[t]{2mm}{\multirow{4}{*}{\rotatebox[origin=c]{90}{\footnotesize{Dense Input}}}}
			& \small{ONet}    &         0.704 &           0.0100 &         0.874 &          0.786\\ 
			& \small{ConvONet}&         0.752 &            0.0074 &         0.894 &          0.881\\ 
			& \small{IF-Net}  &         0.883 &            0.0034 &         0.948 &          0.983\\
			& \small{Ours}    & \textbf{0.958} &   \textbf{0.0024} & \textbf{0.972} & \textbf{0.991}\\
			\hline
\end{tabular}

	\end{center}
	\caption{\textbf{Quantitative Results on $\mathcal{D}_A$}}
	\label{tab:res_ifData}
\end{table}

For this experiment all models are trained and evaluated on $\mathcal{D}_A$, which is noise-free.
The results presented in table~\ref{tab:res_ifData} clearly show that AIR-Nets outperform all baselines by a large margin. Especially in the sparse setting AIR-Nets are dominant. They even outperform all baselines from the dense setting.
We believe that full self-attention in AIR-Nets especially helps in the sparse setting to exploit global symmetries and long-range dependencies, which is less relevant in the dense setting due to high-quality observations. 
Note that in the noise-free scenario, the discretization procedures in ConvONets and IF-Nets have a more relevant negative effect.
Also, note that ConvONets overfitted in both settings before reaching their full potential. 
The qualitative results presented in figure~\ref{fig:recs_DA} confirm the great effectiveness of AIR-Nets. On sparse input data AIR-Nets are capable to reconstruct a surprising amount of detail, while maintaining smooth surfaces. While IF-Nets also produce pleasing results in the dense setting, their reconstruction suffers from bumpy artefacts and often misses fine details. More qualitative results are provided in the supplementary.

\input{figures/onetDataResults}

\subsubsection{Noisy Observations from $\mathcal{D}_B$}
\label{sec:rec_d_b}

In this section, all models are trained and evaluated on $\mathcal{D}_B$, to test the model's performance in the presence of noise.
\begin{table}[t]
	\begin{center}
		\begin{tabular}{|c|l|c|c|c|c|} \hline
			&\multicolumn{1}{l|}{Model}&
			\multicolumn{1}{c|}{IoU$\uparrow$}&
			\multicolumn{1}{c|}{$L_1\text{-CD}$$\downarrow$}&
			\multicolumn{1}{c|}{\small{NC$\uparrow$}}&
			\multicolumn{1}{c|}{\small{F-Score$\uparrow$}} \\ \hline
			\parbox[t]{2mm}{\multirow{4}{*}{\rotatebox[origin=c]{90}{\footnotesize{Sparse Input}}}}
			& \small{ONet}     &         0.755 &          0.0084 &         0.891 &           0.797\\ 
			& \small{ConvONet} &         0.801 &          0.0061 &         0.908 &           0.871\\ 
			& \small{IF-Net}   &         0.763 &          0.0070 &         0.895 &           0.84\\ 
			& \small{Ours}     & \textbf{0.869} & \textbf{0.0047}& \textbf{0.929}  & \textbf{0.93}\\
			 \hline
			 \parbox[t]{2mm}{\multirow{4}{*}{\rotatebox[origin=c]{90}{\footnotesize{Dense Input}}}}
			& \small{ONet}    &         0.758 &            0.0082 &          0.893 &          0.802\\ 
			& \small{ConvONet}&         0.888 &            0.0040 &          0.935 &          0.953\\ 
			& \small{IF-Net}  &         0.924 &    \textbf{0.0030} &          0.951 &  \textbf{0.985}\\
			& \small{Ours}    & \textbf{0.925} &           0.0033 & \textbf{0.952} &         0.978\\
			\hline
\end{tabular}
	\end{center}
	\caption{\textbf{Quantitative Results on $\mathcal{D}_B$}} 
	\label{tab:res_oData}
\end{table}
Table \ref{tab:res_oData} confirms the dominance of AIR-Nets in the sparse setting. In the dense setting, IF-Nets are performing similarly well. Partly, this can be explained due to the fact that IF-Nets are less effected by the noise, since they discretize the input. The qualitative results in figure~\ref{fig:recs_DA} show that AIR-Nets are still capable of producing smooth surfaces, despite the presence of noise. IF-Nets again have a slight tendency to lose thin details, while AIR-Nets tend to reconstruct thin structures too thick (\eg see feet of sofa and the sail).

Another very interesting finding is that all baselines seem to be performing better on~$\mathcal{D}_B$, despite the presence of noise. We believe that the uniformly distributed supervision points of~$\mathcal{D}_B$ could be better suited for all baseline models or that they overfit slower due to the presence of noise.

\subsection{Zero-Shot Generalization}
\label{sec:zero-shot}
In order to evaluate the generalization ability of the models, we evaluate their 3D reconstruction quality on the FAUST \cite{FAUST} dataset in a zero-shot fashion. We chose the best performing model for each model-type trained in the dense setting, i.e. for all baselines the dense models trained on $\mathcal{D}_B$.  For the sake of completeness, we evaluate both dense AIR-Net models. The input point clouds are generated by uniformly sampling 3000 points on the surface of the ground truth mesh, which already contains real measurement noise, as well as occasional holes, violating the watertightness of the objects (\eg see feet of ground truth in figure \ref{fig:zero_shot_faust}). Hence the IoU is calculated using the implicit waterproofing algorithm from~\cite{if-net}. Quantitative and qualitative results are presented in table~\ref{tab:zero_shot_faust} and figure~\ref{fig:zero_shot_faust}, respectively. The results show that IF-Nets and AIR-Nets generalize best, which we account for by their translation equivariance. AIR-Nets trained on $\mathcal{D}_A$, however, reconstruct significantly more details.

\input{figures/faustResults}

\begin{table}[ht]
	\begin{center}
		\begin{tabular}{|l|c|c|c|c|} \hline
			Model&
			{IoU$\uparrow$}&
			{$L_1\text{-CD}$$\downarrow$}&
			{\small{NC$\uparrow$}}&
			{\small{F-Score$\uparrow$}} \\ \hline
			 \small{$\text{ONet}_{\mathcal{D}_B}$}     &         0.149 &          0.05416 &         0.591 &           0.149\\ 
			 \small{$\text{ConvONet}_{\mathcal{D}_B}$} &         0.634 &          0.01066 &         0.816 &           0.661\\ 
			 \small{$\text{IF-Net}_{\mathcal{D}_B}$}   &         0.888 &          0.00354 &         0.942 &           0.958\\ 
			 \small{$\text{ours}_{\mathcal{D}_B}$}     &         0.891 &          0.00353 &         0.942 &           0.957\\ 
			 \small{$\text{ours}_{\mathcal{D}_A}$}     & \textbf{0.920}&  \textbf{0.00290}& \textbf{0.951}&   \textbf{0.964}\\
			\hline
\end{tabular}
	\end{center}
	\caption{\textbf{Quantitative Results of the Zero-Shot Reconstructions on the FAUST dataset.}  
	} 
	\label{tab:zero_shot_faust}
\end{table}

\subsection{Ablation Studies}
\label{sec:ablation_study}

This section presents experiments to validate the design of the \emph{downsampling stage} and \emph{full-attention stage} (see figure \ref{fig:encoder}), as well as, the proposed decoder. For this purpose we construct the most simple baseline composed of a PointNet++ \cite{PointNet++} inspired encoder, no full-attention and interpolation based decoder, as used as a baseline in \cite{ConvOccNets}. We consecutively improve the model by replacing the interpolation-based decoder, with the proposed attention-based decoder, by adding full-attention layers and by using the proposed downsampling stage. The results for these four models are presented in table \ref{tab:component_ablation}, indicating the benefits of each individual module. Additionally, we included a model  using the common maxpooling-based set abstraction instead of the proposed attentive set abstraction module, showing that the proposed set abstraction method compares favourably.

Finally, we compare the proposed decoder against two alternatives in the bottom two rows of table \ref{tab:component_ablation}, showing the great importance of using an expressive decoder. 
More details and experiments are provided in the supplementary.

\begin{table}[ht]
	\begin{center}
		\begin{tabular}{|l|c|c|c|c|} \hline
			Model&
			{IoU$\uparrow$}&
			{$L_1\text{-CD}$$\downarrow$}&
			{\small{NC$\uparrow$}}&
			{\small{F-Score$\uparrow$}} \\ \hline
			 \small{PN::interp}      &         0.444 &          0.03959 &         0.763 &           0.336\\ 
			 \small{PN::ours}        &         0.807 &          0.00528 &         0.891 &           0.890\\ 
			 \small{PN:3full:ours}   &         0.859 &          0.00430 &         0.926 &           0.930\\ 
			 \small{PT:3full:ours}   &         0.888 &          0.00374 &         0.940 &           0.952\\ 
			 \small{ours:3full:ours} & \textbf{0.896}&  \textbf{0.00356}& \textbf{0.945}&   \textbf{0.957}\\ \hline
			 \small{ours:3full:interp}   &     0.801 &          0.00770 &         0.904 &           0.874\\ 
			 \small{ours:3full:LDIF}   &       0.839 &          0.00501 &         0.917 &           0.912\\ 
			\hline
\end{tabular}

	\end{center}
	\caption{\textbf{Component Ablation:} Models are named as composition of 3 sections, the encoder, potential full attention and the decoder. PN and PT stand for PointNet++ and Point Transformer respectively. An empty middle section denotes the lack of full self-attention layers, \emph{interp} denotes a decoder interpolating between features \cite{ConvOccNets} and \emph{LDIF} denotes a decoder interpolating between implicit functions \cite{LDIF}. Note, that the 3rd to last row is the complete AIR-Net. All models were trained on $\mathcal{D}_A$ in the sparse setting for 300 epochs and evaluated on 2000 test examples.
	} 
	\label{tab:component_ablation}
\end{table}

\section{Discussion and Conclusion}
In this work, we have introduced AIR-Nets, a novel encoder-based local implicit shape learning model operating purely on point clouds. Building on the local attention mechanism from  \cite{PointTransformer}, 
we propose a fully translation equivariant framework for 3D shape reconstruction. 
AIR-Nets represent shapes by a global latent vector and a set of local latent vectors anchored in 3D space. This local, modular representation allows for highly detailed reconstruction and promotes generality.
Our experiments show that AIR-Nets outperform previous state-of-the-art methods in 3D reconstruction on the ShapeNet dataset by a large margin. Furthermore, AIR-Nets generalize well to the FAUST dataset in a zero-shot setting. Ablation studies validate several of our encoder components and show the benefits of the proposed, attention-based decoder.

Finally, the AIR-Net framework offers several exciting avenues for future work. First, since our framework is relatively new compared to 3D CNNs, we believe that it offers great potential for future extensions, as well as, applications to large-scale scenes \cite{ConvOccNets} and  joint implicit appearance modelling, \eg \cite{NeRF, SRN, NSVF}. Second, compared to IF-Nets and ConvONets, the relative sparsity of the latent representation renders AIR-Nets more suitable for a probabilistic, generative setting, which could be useful for single-view 3D reconstruction. Finally, our latent representation is compatible with the object detection model from \cite{DETR}, which could ultimately enable semantic 3D reconstruction.

{
\small
\subheading{Acknowledgments} This work was funded by the Deutsche Forschungsgemeinschaft (DFG, German Research Foundation) in the SFB Transregio 161 'Quantitative Methods for Visual Computing' (Project-ID 251654672) and the Cluster of Excellence 'Centre for the Advanced Study of Collective Behaviour'.
}

{\small
\bibliographystyle{ieee_fullname}
\bibliography{AIR-Nets}
}

\end{document}